# Analysis of Gene Regulatory Networks from Gene Expression Using Graph Neural Networks

Hakan T. Otal[*], Abdulhamit Subasi[*], Furkan Kurt[*], M. Abdullah Canbaz[*], Yasin Uzun[†]

[*]Information Sciences and Technology Dept., College of Emergency Preparedness, Homeland Security & Cybersecurity, State University of New York at Albany, Albany, NY,12222, USA
[†] Penn State College of Medicine, Department of Pediatrics, 600 University Dr, Hershey, PA 17033, USA
e-mails: hotal@albany.edu; fkurt@albany.edu; asubasi@albany.edu; mcanbaz@albany.edu; yuzun@pennstatehealth.psu.edu

**ABSTRACT**

Unraveling the complexities of Gene Regulatory Networks (GRNs) is crucial for understanding cellular processes and disease mechanisms. Traditional computational methods often struggle with the dynamic nature of these networks. This study explores the use of Graph Neural Networks (GNNs), a powerful approach for modeling graph-structured data like GRNs. Utilizing a Graph Attention Network v2 (GATv2), our study presents a novel approach to the construction and interrogation of GRNs, informed by gene expression data and Boolean models derived from literature. The model's adeptness in accurately predicting regulatory interactions and pinpointing key regulators is attributed to advanced attention mechanisms, a hallmark of the GNN framework. These insights suggest that GNNs are primed to revolutionize GRN analysis, addressing traditional limitations and offering richer biological insights. The success of GNNs, as highlighted by our model's reliance on high-quality data, calls for enhanced data collection methods to sustain progress. The integration of GNNs in GRN research is set to pioneer developments in personalized medicine, drug discovery, and our grasp of biological systems, bolstered by the structural analysis of networks for improved node and edge prediction.

**Keywords:** Gene Regulatory Networks (GRNs), Graph Neural Networks (GNNs), Graph Attention Network v2 (GATv2), Gene Expression, Single-cell RNA sequencing.

1. **INTRODUCTION**

In the realm of systems biology, understanding the intricate web of interactions within Gene Regulatory Networks (GRNs) stands as a formidable challenge, pivotal for unraveling the complexities of cellular processes, disease mechanisms, and developmental biology. GRNs, which depict the regulatory relationships between genes and their transcription factors, embody a complex system where nodes (genes, proteins) and edges (interactions) form a dynamic and intricate network. Traditional computational methods, while providing significant insights, often struggle with the scale, complexity, and dynamic nature of these networks, leading to a critical need for more sophisticated and adaptable analytical tools. GRN reconstruction using large-scale data has emerged as a pivotal challenge in systems biology, gaining increased relevance in recent years. The integration of complex omics data such as the epigenome or microbiome represents current research trends, and a variety of approaches and evaluation metrics are being developed for reliable GRN reconstruction. These methods offer their own advantages and disadvantages depending on the available data and the purpose of the inference, highlighting the importance of appropriate evaluation methods to determine algorithm performance (Delgado and Gómez-Vela, 2019).

The primary challenge in studying GRNs lies in accurately modeling and predicting the behavior of genes and their regulatory interactions in response to various biological conditions and perturbations. Traditional models face limitations in capturing the non-linear and high-dimensional relationships within GRNs, making it difficult to predict gene expression outcomes under unobserved conditions or to identify potential therapeutic targets for complex diseases. Moreover, the static nature of these models does not account for the dynamic changes in gene expression over time or in response to environmental stimuli, leading to a gap in our understanding of cellular responses and disease progression.

On the other hand, Graph Neural Networks (GNNs) emerge as a promising solution to these challenges, offering a powerful framework for directly modeling the graph-structured data inherent in GRNs. GNNs are capable of learning low-dimensional representations of nodes and edges, capturing both their features and the complex interactions between them. By leveraging the strengths of GNNs, researchers can overcome the limitations of traditional computational methods, enabling a deeper and more nuanced understanding of GRNs. This approach holds the promise of significant advancements in personalized medicine, drug discovery, and our overall understanding of biological systems, marking a pivotal step forward in the field of computational biology and genomics. The utilization of Graph Neural Networks (GNNs) in inferring Gene Regulatory Networks (GRNs) has shown promise in recent research, despite the fact that GNNs typically require substantial training data. Comparisons between GNN models, specifically the GATv2 regression model, and traditional GRN inference models, such as those using XGBoost, have revealed significant differences in stability and accuracy. Despite the challenges, the GATv2 link prediction model has shown superior results in some cases, particularly when correct training data was used. This suggests that with more accurate training data, the performance of GNN-based models for GRN inference could potentially be improved (Feng et al., 2023; Ganeshamoorthy et al., 2022; Makino, 2023)

The integration of Graph Neural Networks into the study of Gene Regulatory Networks represents a frontier in systems biology, offering novel insights and methodologies for deciphering the complex regulatory mechanisms that govern cellular life. As we continue to explore the capabilities of GNNs, we stand on the brink of unlocking a new level of understanding in biology, with the potential to drive forward innovations in medicine, biotechnology, and beyond.

## 2. BACKGROUND AND LITERATURE REVIEW

The exploration of Gene Regulatory Networks (GRNs) through advanced computational methods has become a central focus in the field of computational biology. Traditional methods, while providing foundational insights, often fall short of capturing the dynamic and complex nature of these networks. The advent of Graph Neural Networks (GNNs) presents a revolutionary shift in how researchers approach GRN analysis, offering novel methodologies to decode intricate biological systems (Alawad et al., 2023; Chen and Liu, 2022; Feng et al., 2023; Ganeshamoorthy et al., 2022; Keyl et al., 2023; Wang et al., 2021, 2020).

GNNs emerged as a promising solution to these challenges, offering a powerful framework for directly modeling the graph-structured data embedded in GRNs. GNNs are capable of learning low-dimensional representations of nodes and edges, capturing both their features and the complex interactions between them. By leveraging the strengths of GNNs, researchers can overcome the limitations of traditional computational methods, enabling a deeper and more nuanced understanding of GRNs. Despite the promise shown by GNNs, they typically require substantial training data. Comparisons between GNN models, specifically the GATv2 regression model, and traditional GRN inference models, such as those using XGBoost, have revealed significant differences in stability and accuracy. The GATv2 models, within certain contexts, manifest a degree of result variability that may be perceived as stochastic in nature, thereby presenting a challenge in achieving the outcome consistency observed in gradient boosting frameworks like XGBoost, which are known for their robust performance in regression tasks. This variability, possibly stemming from the inherent stochasticity of the model's training process, underscores the necessity for further examination to enhance the predictive stability and reliability of GATv2 implementations (Jiang et al., 2021).

However, optimizing GNN models for the inference of GRNs necessitates specialized strategies that transcend general machine learning enhancement protocols. Paramount to this endeavor is the curation of training datasets comprehensively representative of the intricate topologies and biological functionalities characteristic of GRNs. Such datasets should integrate multimodal data reflective of diverse regulatory interactions, thereby enriching the GNN's exposure to the spectrum of biological network dynamics. Further refinement can be achieved through hyperparameter tuning, tailored to accommodate the nuanced expression patterns and regulatory complexities inherent to gene networks. Incorporation of domain-specific knowledge, such as established biological pathways, into the GNN architecture could guide the initialization of the network's parameters, fostering a more biologically congruent learning process. Collectively, these targeted adjustments are instrumental in leveraging the unique capabilities of GNNs to elucidate the sophisticated architectures and operational intricacies of biological networks (Feng et al., 2023; Ganeshamoorthy et al., 2022; Makino, 2023).

A recent review highlights the critical role of computational methods in GRN reconstruction, emphasizing the necessity of leveraging machine learning for improved network generation and analysis (Procopio et al., 2023; Sinha et al., 2023). Furthermore, it delineates the importance of model optimization and computational approaches for the validation of constructed networks, thereby underscoring the interdisciplinary nature of GRN studies and the need for continuous advancements in genomics to tackle the complexity of genetic information (Delgado and Gómez-Vela, 2019).

As most network methods rely on gene expression data, accessing and using gene expression data is important for network biology. Multiple web-based tools have been developed to analyze gene expression datasets in this context. One such tool is GEOexplorer, which is a user-friendly webserver designed to enable scientists to access, integrate, and analyze gene expression data from the Gene Expression Omnibus (GEO) database without the need for programming skills, facilitating interactive and reproducible research. The platform supports a range of analyses and visualizations for microarray and RNA-seq datasets, allowing for easy data exploration, interpretation, and the creation of publication-ready figures, thus democratizing the analysis of gene expression data for researchers (Hunt et al., 2022).

Furthermore, in the realm of computational biology, knowledge graphs (KGs) emerge as a valuable resource. The KGs, which are constructed from diverse datasets, serve as a versatile tool, enabling a multitude of applications ranging from semantic data integration to hypothesis generation and decision support in complex biological inquiries. For example, PrimeKG represents a significant advancement in the field of precision medicine by offering a comprehensive multimodal knowledge graph that integrates disparate data sources into a unified framework. This allows for a more nuanced understanding of the multifaceted relationships between genetic and molecular factors and their phenotypic outcomes. By encompassing a wide range of biological scales and including unique drug-disease relationships, PrimeKG facilitates the development of personalized diagnostic and treatment strategies, thereby addressing the complexities inherent in tailoring medical care to individual patient profiles (Chandak et al., 2023).

The integration of Graph Neural Networks into the study of Gene Regulatory Networks represents a frontier in systems biology, offering novel insights and methodologies for deciphering the complex regulatory mechanisms that govern cellular life. As we continue to explore the capabilities of GNNs, we stand on the brink of unlocking a new level of understanding in biology, with the potential to drive forward innovations in medicine, biotechnology, and beyond.

## 3. METHODOLOGY

### 3.1. Data Source

The comprehensive dataset employed in this research was devised for a systematic appraisal of contemporary algorithms purposed for deducing gene regulatory networks from single-cell transcriptional information. Ground truths established for accuracy assessments encompass synthetic networks with discernible developmental trajectories, Boolean models derived from scholarly literature, and a variety of transcriptional regulatory networks. These networks incorporate published Boolean models for discrete biological processes such as mammalian cortical development and the differentiation of the ventral spinal cord, which have been meticulously formulated from scientific literature by harnessing expert knowledge in constructing gene regulatory network Boolean models. These models encapsulate the present understanding of gene interactions and the regulatory dynamics during specific biological processes.

The principal dataset utilized in this analysis comprises a literature-based Boolean dataset for hematopoietic stem cell differentiation (Pratapa et al., 2020). Within this context, the Boolean models act as a benchmark for gauging the accuracy of gene regulatory network inference algorithms and various graph metrics. The researchers adopted the BoolODE system to execute 2,000 simulations, randomly selecting a single cell from each simulation. This procedure was replicated across ten different sampled parameter sets, thus assembling ten distinct datasets. Ultimately, the hematopoietic stem cell dataset encompasses 30 expression samples and reference networks distributed among 2,000 simulations: 10 datasets are without dropouts, 10 possess a moderate dropout rate of q = 50, and 10 exhibit a higher dropout rate of q = 70. This careful approach facilitates a robust evaluation of the algorithms in question across a spectrum of data completeness scenarios.

### 3.2. Gene Regulatory Networks

Cells coordinate their activities by regulating gene transcription in response to signals both inside and outside the cell. Transcription, primarily controlled by transcription factors (TFs), entails proteins that interact with specific DNA sequences (known as DNA binding sites) and influence the transcriptional rate of target genes positively or negatively. The genomic DNA is densely packed with structural proteins into nucleosome complexes, forming the fundamental unit of chromatin, which renders most genes inaccessible to the transcription machinery. For transcription to occur, the promoter region near the transcription start site of a gene must be exposed by displacing nucleosomes. This accessibility change is often initiated by pioneer TFs. Additionally, other TFs can bind to distant cis-regulatory elements (CREs) on the DNA and, in collaboration with cofactors and other proteins, facilitate the recruitment and stabilization of the RNA polymerase protein complex responsible for mRNA synthesis from the gene's DNA sequence (Badia-I-Mompel et al., 2023).

Gene regulatory networks (GRNs) are computational models that represent gene expression regulation as networks, also known as graphs in mathematical terms. GRNs can encompass various components of gene regulation, including transcription factors (TFs), splicing factors, long non-coding RNAs, microRNAs, and metabolites. In this chapter, we focus on the simplest representation of GRNs, which captures the interactions between TFs and target genes and is also subclassified as transcriptional regulatory networks (TRNs) (Uzun, 2023). In this representation, genes (including some TFs) are depicted as nodes, and regulatory interactions between genes are depicted as edges in the GRN. Other types of GRN representations are explored elsewhere. Understanding the topology and dynamics of GRNs is crucial for deciphering how cellular identity is established and maintained, which holds significant implications for engineering cell fate and disease prevention (Badia-I-Mompel et al., 2023).

The quest to understand GRNs has a rich history in biology, exemplified by seminal work in characterizing the bacterial lactose operon from the 1960s. Building large-scale GRNs has become a focal point of systems biology, utilizing various high-throughput experimental techniques and computational algorithms. Historically, GRNs have been constructed from experimentally validated regulatory events compiled in databases or inferred from gene co-expression patterns in bulk transcriptomics data. While transcriptomics data can provide contextualized GRNs tailored to specific biological questions, they may not capture various underlying regulatory mechanisms comprehensively, such as TF protein abundance, DNA binding events, TF cooperation with cofactors, alternative transcript splicing, post-translational protein modifications, and genome accessibility and structure. Incorporating and measuring these additional aspects of gene regulation has the potential to yield GRNs that better reflect gene regulation in vivo. For instance, integrating chromatin accessibility data can refine TF-gene links by considering gene openness and including cis-regulatory elements (CREs) in GRN inference (Badia-I-Mompel et al., 2023).

The intricate interplay among chromatin, transcription factors, and genes gives rise to intricate regulatory circuits, which can be conceptualized as gene regulatory networks (GRNs). Investigating GRNs is crucial for comprehending how cellular identity is established, maintained, and perturbed in diseases. These networks can be deduced from experimental data, historically derived from bulk omics data, or from literature mining. The emergence of single-cell multi-omics technologies has spurred the development of innovative computational approaches that exploit genomic, transcriptomic, and chromatin accessibility data to infer GRNs with unprecedented precision. In this chapter, we discuss the challenges associated with GRN inference, particularly concerning benchmarking, and propose potential avenues for future advancements by incorporating additional data modalities (Badia-I-Mompel et al., 2023).

Moreover, bulk profiling offers aggregated measures across cell types in a tissue sample, which may obscure regulatory programs specific to particular cell types or states. This challenge has been addressed with the advent of single-cell technologies, enabling the inference of GRNs across different cell types, differentiation trajectories, and conditions. Consequently, there has been a surge in novel GRN inference methods, particularly with the introduction of multimodal profiling technologies (Badia-I-Mompel et al., 2023; Duren et al., 2021).

Inferring gene regulatory networks (GRNs) represents a fundamental task in systems biology, seeking to elucidate the intricate connections between genes and their regulators. Unraveling these networks is pivotal for comprehending the intricate regulatory interplay underlying numerous cellular processes and diseases. The emergence of advanced sequencing technologies has facilitated the creation of cutting-edge GRN inference techniques that leverage matched single-cell multi-omics data. Through the utilization of diverse mathematical and statistical approaches, these methods strive to reconstruct GRNs with increased comprehensiveness and precision (Kim et al., 2023).

### 3.3. Graph Neural Networks

Graphs are representations of data where the connections between entities are depicted as edges, while the entities themselves are depicted as nodes (or vertices). In graph structures, the connections and relationships within the data are fundamental components and play a crucial role in how we conceptualize the world, analyze data, and derive insights from it. This stands in contrast to the structured, grid-like data typically encountered in traditional learning and analytics scenarios, such as relational database tables, pandas dataframes, or spreadsheets. In such grid-like structures, the relationships between data points are less pronounced compared to the more explicit networked or graph-based representations (Keita Broadwater and Namid Stillman, n.d.) In fields saturated with enthusiasm for emerging technologies and methodologies, Graph Neural Networks (GNNs) stand out as a significant advancement for both deep learning and graph analytics. Previously, integrating graph features into model training in deep learning was cumbersome and lacked scalability, failing to seamlessly incorporate node and edge properties. Similarly, despite the existence of various methods in graph analysis for characterizing networks and their elements, many of which are utilized in modern graph databases and analytical software, they have struggled to fully incorporate node and edge properties. Moreover, developing generalizable models applicable to unseen nodes, edges, or graphs has been challenging until now (Keita Broadwater and Namid Stillman, n.d.).

Graph neural networks (GNNs) represent a category of methods aimed at extending and customizing deep learning models to effectively train on and learn from data structured as graphs. Broadly termed as "Graph Intelligence," this field encompasses AI techniques tailored for analyzing graph data. Unlike conventional deep learning models, which struggle with non-grid graph topologies, GNNs are specifically designed to handle such structures. The origins of graph intelligence are rooted in the pioneering introduction of the first GNN two decades past, specifically designed for analytical tasks associated with graphs and their constituent nodes. In the ensuing years, the ascendance of deep learning as a transformative force across a multitude of disciplines has catalyzed concerted efforts to propel the evolution of graph intelligence methodologies and their myriad applications. It is now broadly recognized that deep learning constitutes the bedrock upon which the edifice of graph intelligence is constructed (Abdel-Basset et al., 2023).

A graph neural network (GNN) is an algorithm designed to interpret and learn from graphs, encompassing their nodes, edges, and associated features. GNNs offer similar benefits to conventional neural networks but are tailored to operate on graph data. Traditional machine learning and deep learning techniques encounter difficulties when applied to graph structures. Representing graph data in grid-like formats and traditional data structures poses challenges, including issues such as permutation invariance, where the ordering of input graph representations can affect the learning process. Moreover, traditional methods often overlook the underlying network structure during learning (Keita Broadwater and Namid Stillman, n.d.). Graph neural networks (GNNs) have proven to be highly effective in learning over graphs across various application domains. To address the challenge of scaling up GNN training for large and ever-expanding graphs, distributed training emerges as the most promising solution (Lin et al., 2023).

GNNs have emerged as potent tools for learning from graph-structured data across diverse domains, including recommendation systems, question-answering, drug discovery, and astrophysical simulations. However, beyond their task performance, other factors such as susceptibility to adversarial attacks, potential biases, and resource demands in edge computing environments need consideration. While performance-focused GNNs offer notable benefits, they may also present challenges such as vulnerability to attacks, unfair treatment of specific groups, or excessive resource consumption in edge computing settings (Sharma et al., 2023; Wu et al., 2021; Zhang et al., 2024).

GNNs represent a promising field within machine learning aimed at tackling real-world challenges across various domains, such as social networks, recommender systems, computer vision, and pattern recognition. A crucial element of GNNs is their operators, responsible for training on graph-structured data and transmitting learned node information to subsequent layers. While GNNs excel in capturing complex interactions and are extensively used in personalized tasks like recommendation systems, traditional personalization methods often rely on centralized GNN learning on global graphs, posing significant privacy risks due to the sensitive nature of user data (Wu et al., 2022).

Using a Graph Neural Network involves an additional step in the process. Alongside initializing the neural network parameters, we also initialize a representation of the graph nodes. Thus, in our iterative process:

1. *Input and Node Representation:*
   - The procedure initiates with the input of graph-structured data.
   - Subsequent to input, an update of node representations is conducted via Graph Neural Network (GNN) layers, which encapsulate the relational dependencies among nodes.
2. *Data Processing through Neural Network Layers:*
   - The data, characterized by newly computed node representations, is subsequently propagated through traditional neural network layers. This phase encompasses the application of learned filters and activation functions to distill features and facilitate forward propagation.
3. *Prediction and Iterative Update:*
   - The final step involves the generation of predictive outputs alongside the updated node representations.

The iterative refinement of neural network parameters is augmented by GNN layers, which are adeptly configured to analyze the inherent structure of graphs. These layers operate by enabling a pattern of interaction that extends to nodes within a specified 'x' hop distance, integrating both local and broader topological features of the graph. It is important to note that GNN layers typically avoid the profound layering characteristic of traditional deep learning models. In the domain of GNNs, a pattern emerges where the benefits of adding numerous layers begin to diminish, suggesting there is an optimal level of depth that maximizes performance without incurring unnecessary complexity. This observation highlights the delicate equilibrium in GNN design that must be achieved to effectively leverage their analytical capabilities in graph-based data (Keita Broadwater and Namid Stillman, n.d.).

## 4. RESULTS AND DISCUSSION

In computational biology, the exploration of Gene regulatory networks (GRNs) is rapidly growing. Given the complexity and scale of these networks, many researchers employ machine learning techniques to deduce GRNs from gene expression data, often obtained through RNA-seq analysis. Uncovering the regulatory relationships among genes and reconstructing GRNs based on gene expression data represent fundamental challenges in bioinformatics. The task of computationally inferring potential regulatory connections between genes can be framed as a link prediction problem within a graph structure. Leveraging Graph neural network (GNN) methodologies offers an avenue to construct GRNs by leveraging the propagation of information among neighboring nodes throughout the entire gene network. Figure 4 illustrates a general framework of gene regulatory graph neural network (GRGNN) to build GRNs from scratch using single cell RNA sequence.

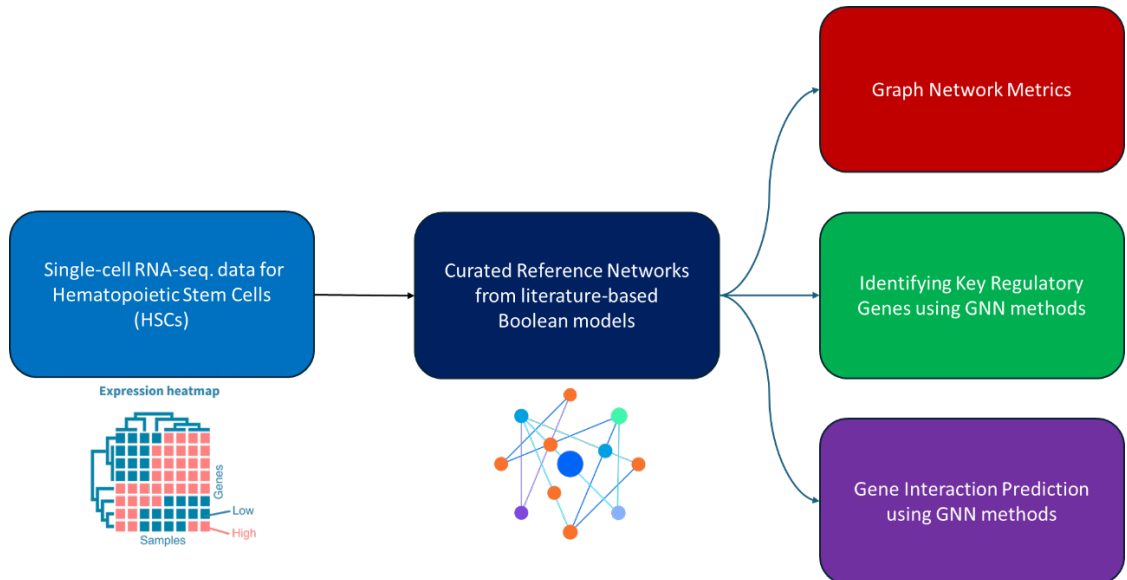

**Figure 1:** General Framework of general framework of gene regulatory graph neural network.

### 4.1. GRN Construction

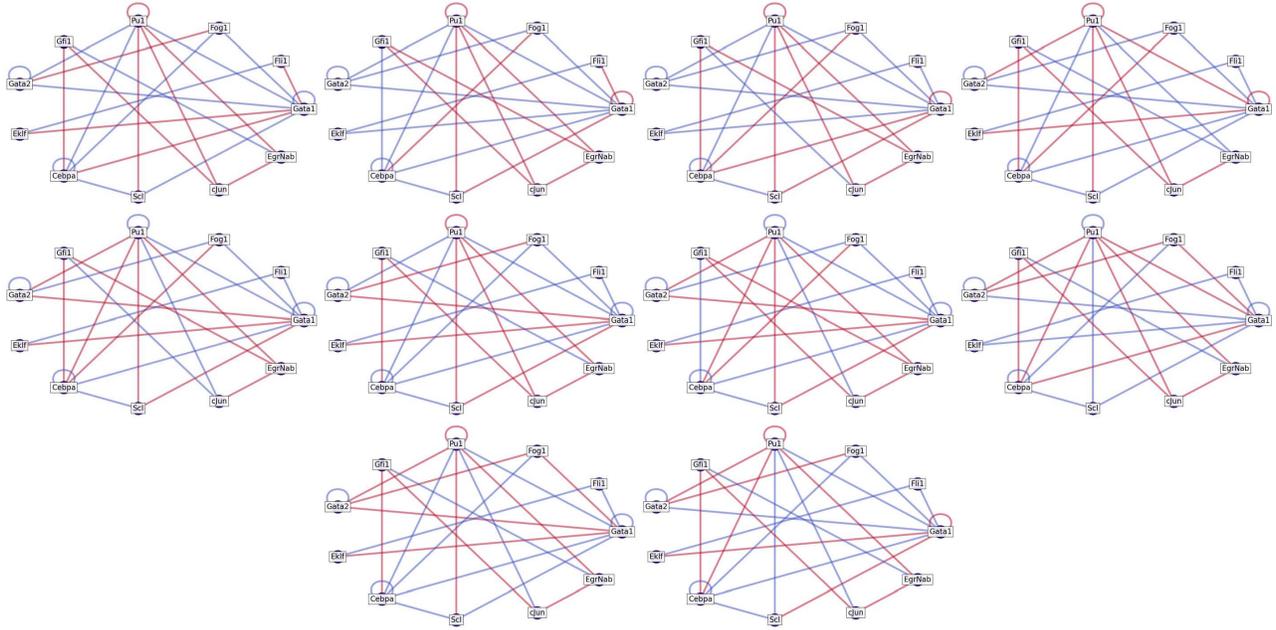

**Figure 2**: Ten Sample reference networks with q=50 dropout from the HSC dataset.

The construction of Gene Regulatory Networks (GRNs) using Graph Neural Networks (GNNs) starts with the assembly of a comprehensive dataset that includes transcriptional data from various sources. This dataset often comprises gene expression profiles from single-cell RNA sequencing, highlighting the dynamic nature of gene regulation across different cellular conditions. The primary step in GRN construction involves the creation of a graph where nodes represent genes or transcription factors, and edges denote regulatory interactions, either activation or repression. The strength and direction of these interactions are initially unknown and are to be inferred through computational models.

To construct a GRN, the dataset undergoes preprocessing to normalize gene expression levels, mitigate batch effects, and identify significantly varying genes. This refined dataset, as explained in *Section 3.1 - Data Source*, serves as the input for GNN models. GNNs, such as Graph Attention Networks (GATv2), are trained to recognize patterns in the data that signify regulatory relationships. These models leverage the inherent graph structure of GRNs, enabling the capture of complex interaction dynamics through node features (gene expression levels) and edge features (potential regulatory effects). In Figure 2, we display 10 sample reference networks with q=50 dropout from the HSC dataset.

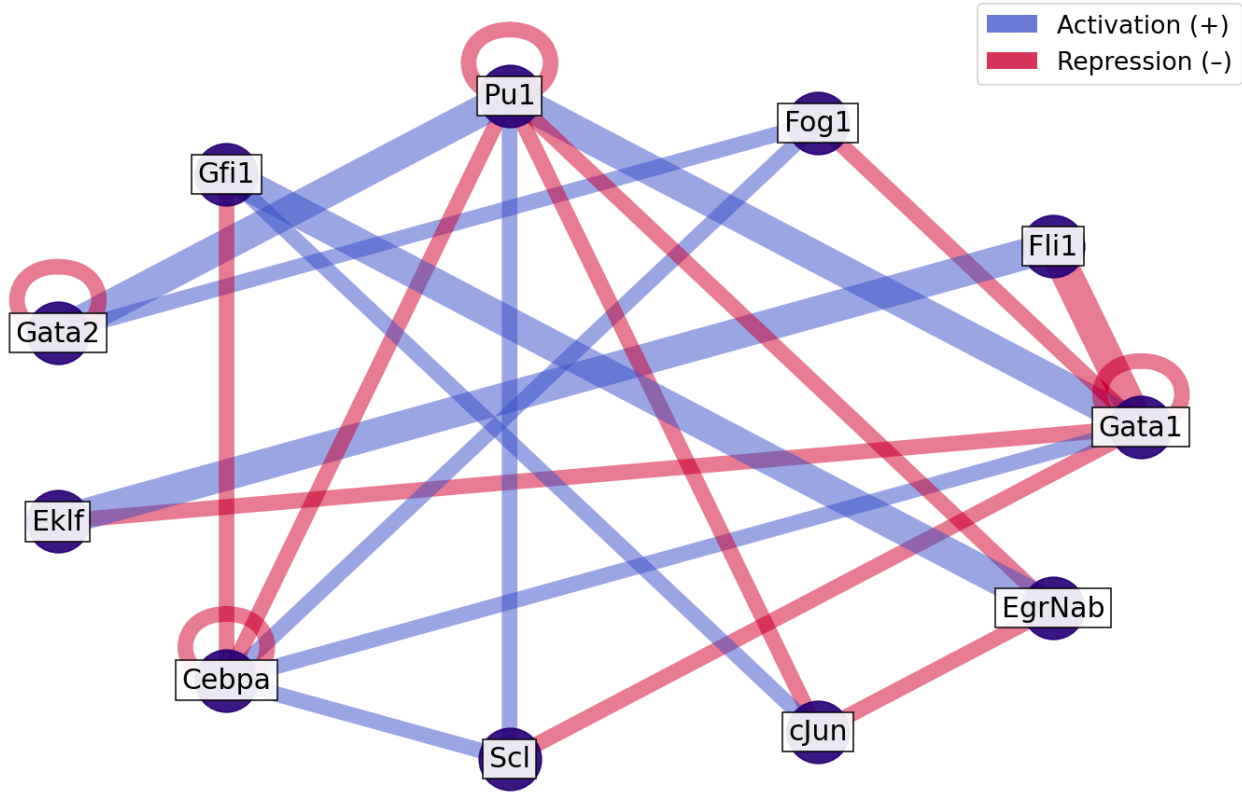

**Figure 3**: Gene Regulatory Networks for HSC (Aggregated Eight Samples from 2000 Sim. by calculating the summations of their link weights, Dropouts = 50%)

In Figure 3, we represent the aggregated gene regulatory network (Aggregated-GRN) by calculating the summations of their link weights (+1 for each activation, -1 for each repression) for hematopoietic stem cells (HSCs) shown in Figure 1. Aggregated-GRN shows the interactions between various transcription factors (TFs) that are known to play roles in the regulation of gene expression within these cells. The transcription factors are represented as nodes in the network, with their names (e.g., Gata1, Pu.1, Fli1) displayed inside boxes. The edges connecting these nodes represent regulatory interactions, with blue lines indicating activation (positive regulation) and red lines indicating repression (negative regulation).

**Central regulators** like *Pu.1, Gata1*, and *Cebpa* are pivotal, as they form numerous activation and repression connections, making them key to HSC behavior. Such transcription factors have the capacity to fine-tune cellular responses through their dual roles in activating and repressing genes.

The network reveals **feedback loops**, where a TF regulates another that, in turn, impacts the first—this can lead to stable cellular states crucial for fate decisions. There's an evident **redundancy and robustness** in the system, where multiple factors target the same downstream TF, allowing the network to maintain its function even if one pathway is disturbed. The thickness of the lines suggests the strength of the regulatory interactions, with thicker lines indicating more consistent or strong influences, which could point to biologically significant pathways. Moreover, a potential **hierarchical structure** is implied by the directionality and number of connections, indicating that some TFs could act as master regulators. However, the network's 50% dropout rate signals a significant uncertainty, hinting at the stochastic nature of gene regulation or experimental limitations.

In some cases, maintaining a balance between activation and repression is critical for sustaining HSC identity; disruptions here could lead to conditions like leukemia. The network's major hubs represent **candidate genes for further study**; these genes are integral for understanding stem cell maintenance and differentiation. Moreover, this network embodies the regulatory mechanisms underpinning cellular plasticity and differentiation—knowledge that is invaluable in regenerative medicine, where directing stem cell differentiation into specific cell types is the goal. Manipulating key transcription factors could thus unlock new therapeutic strategies.

Overall, this network provides a visual summary of the regulatory interactions among different transcription factors that control the gene expression programs within hematopoietic stem cells. Understanding these interactions is crucial for grasping the complex control mechanisms that determine cell fate and function, especially in the context of blood cell development and differentiation. To support open science and future endeavors, we provide our source code on GitHub to the larger community[1].

### 4.2. Graph Structure Analysis

Once a GRN is constructed, the next phase involves the analysis of its graph structure to understand the network's topology by looking at the centrality metrics, identifying key regulatory pathways, and discerning the characteristics of gene interactions. Graph structure analysis encompasses several metrics, including node degree distribution, which reflects the connectivity of genes within the network; clustering coefficients, indicating the tendency of genes to form tightly knit groups; and various centrality metrics, which measure the closeness between gene pairs in regulatory pathways.

In Table 1, we present the various network metrics of the nodes of the aggregated GRN, shown in Figure 2. Additionally, In Figure 3, we display these network metrics, centrality degrees, for the aggregated GRN.

**Table 1:** The network metrics of the nodes of the aggregated GRN shown in Figure 3.

---

[1] https://github.com/AI-in-Complex-Systems-Lab/HSC-GRN-GNN

|  | Clustering Coefficients | Degree Centrality | Closeness Centrality | Betweenness Centrality | Eigenvector Centrality |
| --- | --- | --- | --- | --- | --- |
| Gata1 | 0.333 | **0.900** | **0.769** | **0.408** | **0.507** |
| Fli1 | **1.0** | 0.200 | 0.476 | 0.000 | 0.129 |
| Fog1 | 0.667 | 0.300 | 0.526 | 0.012 | 0.252 |
| Pu1 | 0.333 | **0.800** | **0.714** | **0.310** | **0.468** |
| Gfi1 | 0.333 | 0.300 | 0.500 | 0.033 | 0.149 |
| Gata2 | 0.667 | 0.500 | 0.556 | 0.019 | 0.312 |
| Eklf | **1.0** | 0.200 | 0.476 | 0.000 | 0.129 |
| Cebpa | 0.400 | **0.700** | **0.667** | 0.173 | **0.422** |
| Scl | **1.0** | 0.300 | 0.588 | 0.000 | 0.283 |
| cJun | 0.667 | 0.300 | 0.500 | 0.012 | 0.157 |
| EgrNab | 0.667 | 0.300 | 0.500 | 0.012 | 0.157 |

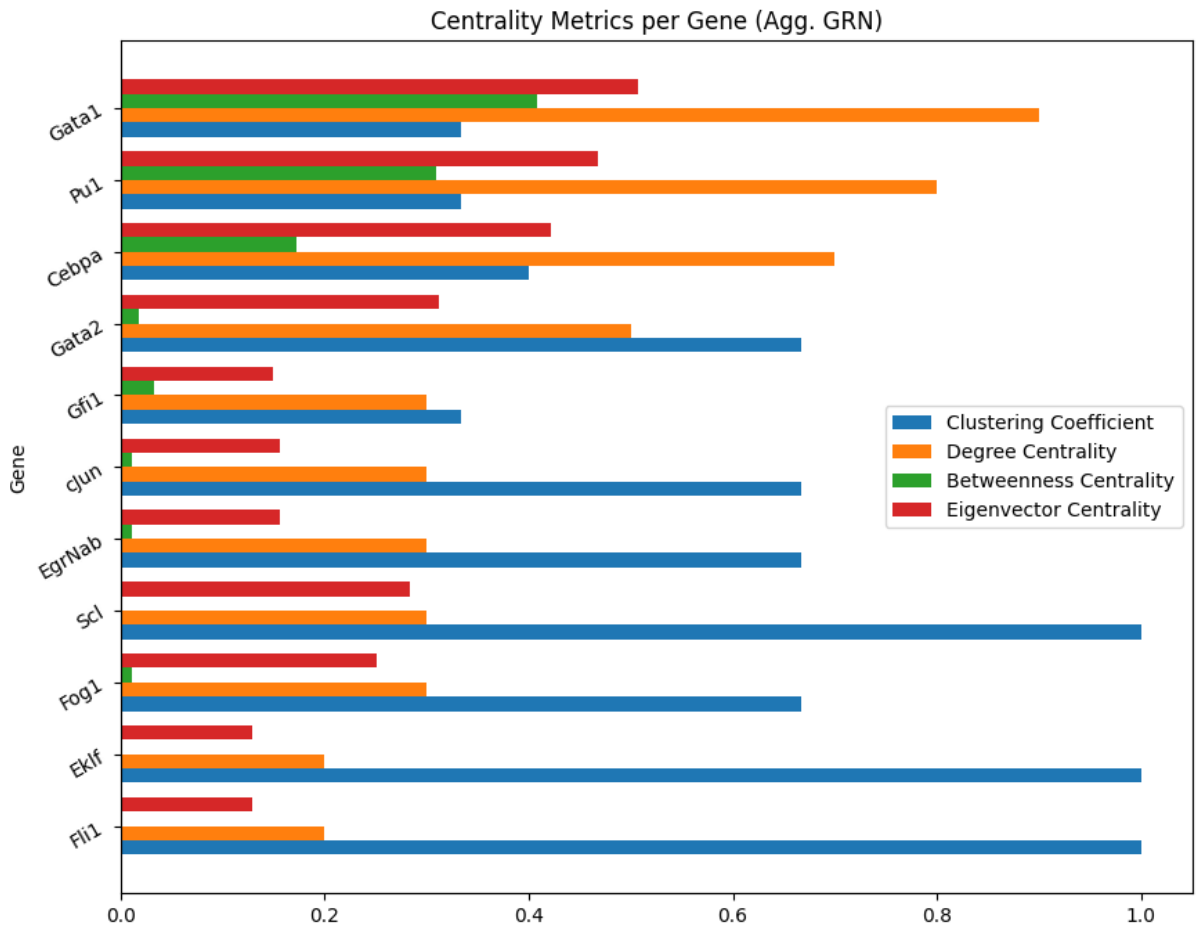

**Figure 4**: The centrality degrees of the nodes in the gene regulatory network shown in Figure 2.

In the analysis of the gene regulatory network's centrality measures, several key observations emerge, highlighting the intricate dynamics of transcription factor interactions.

**Clustering Coefficient** is instrumental in understanding the tendency of transcription factors to form cohesive groups. Notably, *Fli1, Eklf*, and *Scl*, each with a coefficient of **1.0**, display strong interconnectivity, suggesting they reside within highly integrated modules. This arrangement implies a network structure designed for redundancy, potentially enhancing the network's resilience to external disturbances.

**Degree Centrality** is another critical indicator of a transcription factor's connectivity within the network. *Gata1* stands out with the highest degree centrality of **0.900**, underscoring its extensive interactions with numerous other transcription factors. This central placement indicates Gata1's pivotal role in orchestrating a wide array of genes pivotal for HSC function, and it may be considered a primary regulatory hub within the network.

Moving to **Closeness Centrality**, *Gata1* again manifests a significant role, evidenced by a high value of **0.769**. This metric denotes *Gata1's* proximity to other nodes in the network, suggesting its capacity for rapid influence across the gene regulatory landscape, which is essential for the prompt response of HSCs to regulatory signals.

In the context of **Betweenness Centrality**, which captures the frequency of a node acting as an intermediary within the shortest communication paths, *Gata1* is observed with a noteworthy value of **0.408**. This prominence accentuates its function as an information conduit within the network, vital for maintaining the integrity of gene regulation processes in HSCs.

Lastly, **Eigenvector Centrality** provides insight into the influence of a node relative to the connectivity of its neighbors. Once more, *Gata1* attains a considerable score of **0.507**, reinforcing its status as an influential node not merely through its direct connections but also through its strategic associations with other significant transcription factors within the network.

These centrality measures collectively underscore the significance of **Gata1** within the gene regulatory network, illustrating its potential as a focal point for therapeutic intervention and further scientific inquiry into the regulatory mechanisms of hematopoietic stem cells.

## 4.3. Identifying Key Regulators

Identifying key regulators within a GRN is crucial for understanding the mechanisms of gene regulation and their impact on cellular processes. This involves analyzing the influence of individual genes or transcription factors on the network's dynamics. Key regulators are typically characterized by their high connectivity, controlling a significant portion of the network, or by their strategic position, acting as bridges between different functional modules.

GNN models, equipped with attention mechanisms, play a pivotal role in identifying these key regulators. These mechanisms allow the model to assign importance weights to nodes and edges, reflecting the significance of genes and their interactions in regulatory activities. By examining the attention weights, researchers can pinpoint genes that play central roles in the GRN, making them potential targets for therapeutic interventions or further biological investigation.

We used 2 GATConv (Graph attention layer from Graph Attention Network) layers in the model and trained with the sample networks. After training, we can interpret the model's attention weights to understand node importance. For GAT, attention weights can be extracted from the learned parameters.

In Figure 5, we visually quantify the influence of individual genes in a gene regulatory network, employing lengths of bars to convey their relative significance. Notably, **Gata2** emerges as a principal gene, indicated by the longest blue bar, asserting its potential as a critical influence on the network's dynamics. Conversely, other genes manifest with shorter or negatively oriented bars, suggesting a spectrum of positive to negative impacts on the network's function, with implications that certain genes enhance while others may inhibit systemic behavior.

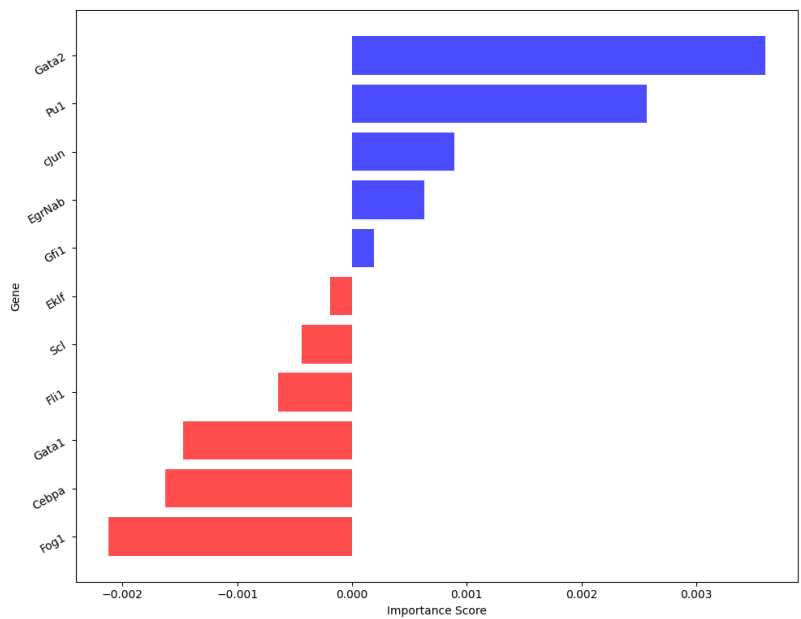

**Figure 5**: Node Importance Scores: the extracted attention weights for each gene in Figure 4.

This analysis, facilitated by Graph Attention Networks (GATs), provides a nuanced interpretation of the model by pinpointing influential nodes and edges. Such insights, revealing the prominence of genes like **Gata2** and **Pu.1**, and the inhibitory roles of genes like **Fog1** and **Cebpa**, are invaluable, proposing candidates for further biological investigation or therapeutic targeting. The interpretability offered by GATs aids in the formulation of testable biological hypotheses, as the model's indication of a gene's role can be empirically evaluated through methods like gene editing. Ultimately, this approach exemplifies the synergy between machine learning and bioinformatics, enhancing our understanding of biological networks and their complex interplay.

### 4.4. Link Prediction

Link prediction in the context of GRNs involves inferring missing interactions or predicting new regulatory relationships between genes. This is particularly challenging due to the complex and dynamic nature of gene regulation. GNN models are well-suited for this task, as they can learn the underlying patterns of gene interactions from the network structure and node features.

Training GNNs for link prediction involves using a subset of known regulatory interactions to learn a representation of the GRN that captures its structural and functional properties. The model can then predict the existence of links between gene pairs that are not present in the training set. Evaluating the performance of these predictions against a set of validated regulatory interactions helps refine the model, improving its accuracy in uncovering novel aspects of gene regulation.

In this experiment, we utilized a Graph Neural Network (GNN) architecture composed of two Graph Attention Network (GATConv) layers. The network was trained using a strategy that incorporated both positive and negative edge oversampling to optimize for predictive performance. The model's evaluation was multi-faceted, encompassing a suite of metrics to assess its predictive capabilities thoroughly. Notably, the test loss registered at a promising **1.68%,** reflecting the model's adeptness at forecasting gene interactions in concordance with the empirical interactions captured in the test dataset. This low test loss implies a high level of accuracy in the model's predictions relative to the true data (Figure 6).

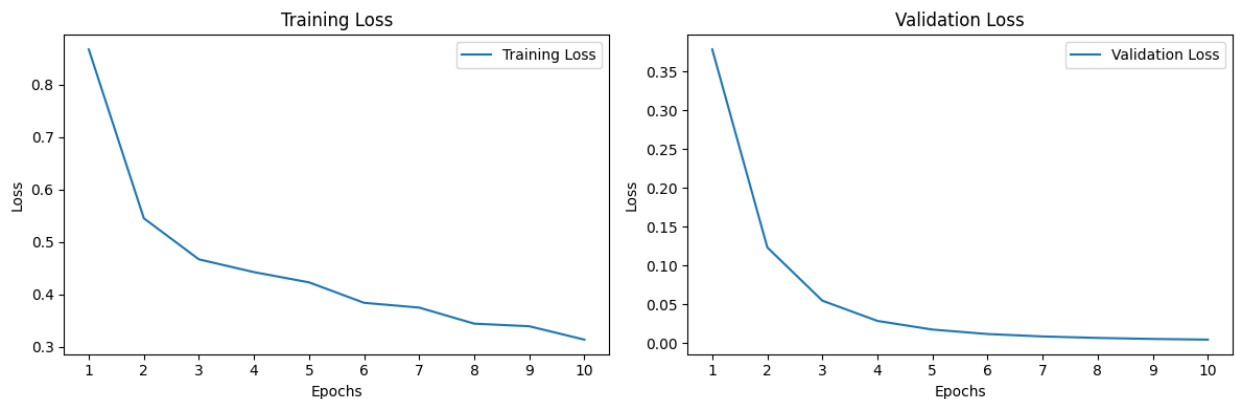

**Figure 6**: Training and Validation Loss of GRN with two GATConv.

Complementing the test loss, the model achieved an impressive test accuracy, thereby successfully predicting gene interactions with **96.09%** correctness. Moreover, the precision of the model was flawless, with a score of **1.0**, indicating that each gene interaction the model predicted to exist was indeed validated in the test set. However, the test recall was slightly lower, signifying that the model identified **92.17%** of all actual gene interactions present in the test data. To provide a singular, balanced metric of performance, the test F1-score was calculated at **95.93%,** which, as the harmonic mean of precision and recall, offers an insightful gauge of the model's overall predictive prowess, particularly in balancing false positives and false negatives.

These results suggest that the GNN model performed well in predicting gene interactions, demonstrating the potential of GNNs in advancing our understanding of GRNs. However, it's important to note that the performance of the model can be influenced by the quality and quantity of the training data. Therefore, efforts to gather more accurate and comprehensive gene interaction data could further improve the performance of GNN-based models for GRN inference. This highlights the importance of continuous advancements in data collection and preprocessing techniques in the field of computational biology.

### 4.5. Discussion

The application of Graph Neural Networks (GNNs) to the analysis of Gene Regulatory Networks (GRNs) has yielded promising results, demonstrating the potential of this approach to revolutionize our understanding of complex biological systems. In this study, we constructed GRNs using a GATv2 model trained on a comprehensive dataset comprising gene expression profiles and literature-curated Boolean models. The model exhibited strong performance in predicting gene interactions, achieving a test accuracy of 96.09%, a precision of 100.0%, a recall of 92.17%, and an F1-score of 95.93%. These metrics indicate the model's ability to accurately identify regulatory relationships between genes, showcasing the power of GNNs in capturing the intricate dynamics of GRNs.

Analyzing the centrality measures of the gene regulatory network has elucidated several key aspects of hematopoietic stem cell regulation. The transcription factor Gata1 emerges as a pivotal regulatory hub, distinguished by its high centrality across various metrics. Specifically, Gata1 exhibits a degree centrality of 0.900, a closeness centrality of 0.769, a betweenness centrality of 0.408, and an eigenvector centrality of 0.507, underscoring its extensive influence on the network. Its central position suggests it is integrally involved in essential pathways for HSC maintenance and lineage specification. Given the centrality of Gata1, it represents a potential target for therapeutic interventions. Modulating its activity could precipitate significant alterations in HSC behavior, offering avenues for the treatment of hematological disorders.

The existence of regulatory subnetworks is hinted at by the high clustering coefficients found in certain transcription factors. For example, Fli1, Eklf, and Scl, all display clustering coefficients of 1.000, indicating a propensity for these nodes to form tightly interconnected modules. These modules may be responsible for discrete regulatory functions or specific cellular states within the broader regulatory landscape of HSCs. Additionally, the betweenness centrality measure reveals a nuanced picture of information flow within the network. While Gata1 acts as a principal conduit, other transcription factors such as Pu1 (betweenness centrality of 0.310) also contribute significantly to the network's integrity. This distribution of control points to a robust network architecture capable of sustaining regulatory functions in the face of perturbations.

These centrality metrics underscore the intricate regulation of HSCs and the critical role network dynamics play in gene expression and cell behavior. These insights pave the way for potential targeted manipulation of these networks, opening new doors in the realm of regenerative medicine by potentially influencing stem cell fate decisions. Such strategies could have profound therapeutic implications, marking a substantial advance in our ability to modulate stem cell-related processes.

The exploration of gene regulatory networks (GRNs) from gene expression data is fundamental in molecular biology, with implications ranging from fundamental biological processes to complex disease mechanisms. While traditional methods for GRN analysis have relied on statistical and mathematical approaches, recent advancements in machine learning, particularly the utilization of graph neural networks (GNNs), offer promising avenues for more comprehensive and accurate analysis. GNNs have distinct advantages for GRN analysis. They can directly operate on graph-structured data, effectively capturing the topology and dynamics of GRNs compared to traditional methods. By representing genes as nodes and regulatory interactions as edges in a graph, GNNs can learn patterns and dependencies from gene expression profiles, enhancing the accuracy of inferred regulatory relationships. Additionally, GNNs excel in capturing non-linear interactions and higher-order dependencies in GRNs, providing insights into complex regulatory mechanisms such as feedback loops and combinatorial regulation.

Moreover, GNNs can integrate multi-modal data sources, including chromatin accessibility and DNA methylation data, to provide a comprehensive view of regulatory interactions within cells. By incorporating diverse data types, GNNs can uncover hidden regulatory relationships, improving the accuracy of inferred GRNs. However, challenges exist in using GNNs for GRN analysis. Interpretability of GNN-based models remains a concern, as the learned representations may not always be easily understandable by biologists. Scalability to large-scale GRNs is also a consideration, as computational and memory requirements may become prohibitive with increasing network complexity.

4.5.1. Implications for Cancer Research in the Asia-Pacific Region

The application of Graph Neural Networks (GNNs) to Gene Regulatory Network (GRN) analysis, as demonstrated in this study, holds significant promise beyond theoretical advancements. It has particular relevance for addressing pressing health concerns in vulnerable regions, such as the Asia-Pacific area, where cancer poses a substantial burden (Youn and Han, 2020).

Cancer represents a significant global health challenge, with underdeveloped countries, particularly those in the Asia-Pacific region, bearing a disproportionate burden of the disease (Mikhail et al., 2022). For instance, the elevated prevalence of somatic TP53 mutations across various cancer types in the Asia-Pacific region underscores the urgency for effective, tailored diagnostic and therapeutic interventions (Ghosh et al., 2022).

The diversity of TP53 mutations, occurring at different stages of tumor development, often complicates the extraction of clinically relevant information (Ghosh et al., 2022). This is where the sophisticated modeling capabilities of GNNs can prove invaluable. By leveraging GNNs to analyze GRNs, researchers can gain a more nuanced understanding of cancer development and progression in the context of TP53 mutations. This approach aligns with our findings on the effectiveness of GNNs in capturing complex regulatory dynamics and identifying key regulators within gene networks (Mahanta et al., 2022; Mikhail et al., 2022).

However, it is crucial to note that the successful implementation of GNNs in this context, as in our study, relies heavily on high-quality data. This underscores the need for robust data collection methodologies, particularly in regions like the Asia-Pacific, to fully leverage the potential of GNNs in revolutionizing cancer research and healthcare. The development of comprehensive, region-specific datasets could significantly enhance the accuracy and relevance of GNN-based models for GRN analysis in cancer research.

The application of our GNN-based approach to cancer-specific GRNs could potentially lead to the identification of novel therapeutic targets or biomarkers specific to the Asia-Pacific population. This could pave the way for more personalized and effective cancer drugs, addressing the unique genetic and environmental factors influencing cancer development in this region (Liu et al., 2024).

## 5. CONCLUSION

The results of this study underscore the significant advantages of using GNNs for GRN analysis. Unlike traditional computational methods, which often struggle with the scale and complexity of GRNs, GNNs excel at modeling the graph-structured data inherent in these networks. By leveraging the strengths of GNNs, researchers can overcome the limitations of traditional approaches, enabling a deeper and more nuanced understanding of gene regulatory mechanisms. However, it is essential to acknowledge that the performance of GNN-based models for GRN inference is heavily influenced by the quality and quantity of the training data. While the results obtained in this study are promising, further improvements in data collection and pre-processing techniques could potentially enhance the accuracy and robustness of these models. Continuous advancements in computational biology, coupled with the integration of diverse data sources, such as epigenomic and microbiome data, will be crucial in refining our understanding of GRNs and their role in cellular processes, disease mechanisms, and developmental biology.

Hence, the integration of Graph Neural Networks into the study of Gene Regulatory Networks represents a ground-breaking frontier in computational biology. By harnessing the power of GNNs, researchers can unravel the intricate web of interactions within these complex networks, unlocking new avenues for personalized medicine, drug discovery, and a deeper comprehension of the fundamental mechanisms that govern cellular life. As we continue to explore the capabilities of GNNs and refine our data collection and analysis techniques, we stand poised to drive forward transformative innovations in biology, medicine, and biotechnology. Besides, the application of graph neural networks to analyze gene regulatory networks offers new opportunities for advancing our understanding of gene regulation. Despite challenges, GNNs provide valuable insights into the complex regulatory dynamics governing gene expression, offering a promising tool for researchers in molecular biology and bioinformatics.